\documentclass[conference]{IEEEtran}
\IEEEoverridecommandlockouts

\usepackage{cite}
\usepackage{amsmath,amssymb,amsfonts}
\usepackage{algorithmic}
\usepackage{graphicx}
\usepackage{textcomp}
\usepackage{xcolor}
\usepackage{makecell,booktabs}
\usepackage{enumitem}
\usepackage{bbm}
\usepackage{mathtools}

\definecolor{refblue}{rgb}{0.21,0.49,0.74}
\usepackage[pagebackref,breaklinks,colorlinks,allcolors=refblue]{hyperref}

\newcommand{\STAB}[1]{\begin{tabular}{@{}c@{}}#1\end{tabular}}
\def\BibTeX{{\rm B\kern-.05em{\sc i\kern-.025em b}\kern-.08em
    T\kern-.1667em\lower.7ex\hbox{E}\kern-.125emX}}
\begin{document}

\title{KARST: Multi-Kernel Kronecker Adaptation with Re-Scaling Transmission for Visual Classification
}

\author{\IEEEauthorblockN{Yue Zhu$^{\star}$}
\IEEEauthorblockA{
\textit{Dalian University of Technology}\\
Dalian, China \\
zhuyuedlut@gmail.com}
\and
\IEEEauthorblockN{Haiwen Diao$^{\star}$}
\IEEEauthorblockA{
\textit{Dalian University of Technology}\\
Dalian, China \\
diaohw@mail.dlut.edu.cn}
\and
\IEEEauthorblockN{Shang Gao$^{\star}$}
\IEEEauthorblockA{
\textit{Dalian University of Technology}\\
Dalian, China \\
gs940601k@mail.dlut.edu.cn}
\and
\IEEEauthorblockN{Long Chen}
\IEEEauthorblockA{
\textit{The Hong Kong University of Science and Technology}\\
Hong Kong, China \\
longchen@ust.hk}
\and
\IEEEauthorblockN{Huchuan Lu$^{\dagger}$\thanks{$^{\star}$Equal contribution. $^{\dagger}${Corresponding author.}}}
\IEEEauthorblockA{
\textit{Dalian University of Technology}\\
Dalian, China \\
lhchuan@dlut.edu.cn}
}

\maketitle

\begin{abstract}
Fine-tuning pre-trained vision models for specific tasks is a common practice in computer vision. However, this process becomes more expensive as models grow larger. 
Recently, parameter-efficient fine-tuning (PEFT) methods have emerged as a popular solution to improve training efficiency and reduce storage needs by tuning additional low-rank modules within pre-trained backbones.
Despite their advantages, they struggle with limited representation capabilities and misalignment with pre-trained intermediate features. 
To address these issues, we introduce an innovative Multi-Kernel Kronecker Adaptation with Re-Scaling Transmission (KARST) for various recognition tasks. 
Specifically, its multi-kernel design extends Kronecker projections horizontally and separates adaptation matrices into multiple complementary spaces, reducing parameter dependency and creating more compact subspaces. Besides, it incorporates extra learnable re-scaling factors to better align with pre-trained feature distributions, allowing for more flexible and balanced feature aggregation.
Extensive experiments validate that our KARST outperforms other PEFT counterparts with a negligible inference cost due to its re-parameterization characteristics.
Code is publicly available at: \href{https://github.com/Lucenova/KARST}{\textcolor{blue}{https://github.com/Lucenova/KARST}}.
\end{abstract}


\section{Introduction}

Recently, large pre-trained foundational models have exhibited strong representation capabilities in various downstream domains~\cite{TransF:ViT,VLM:EVE,ITM:SGRAF}. They are trained on extensive in-domain or web-scale data, enabling them to generalize effectively across various downstream tasks. However, as their model size grows, fully fine-tuning them for specific tasks becomes increasingly expensive and impractical. To address it, parameter-efficient transfer learning (PETL) methods~\cite{TL:UniPT,diao2025sherl} have gained traction, significantly reducing the computational and storage costs of fine-tuning large models. Among them, Partially-tuning approaches~\cite{TL:BitFit} focus on training a subset of pre-trained parameters for downstream domains. Besides, Prompt-based methods~\cite{TL:VPT} attempt to integrate several trainable prompt tokens into model inputs, while Adapter-based studies~\cite{diao2024gssf} try to insert several multi-layer perceptrons (MLPs) into transformer blocks, either after or in parallel with multi-head self-attention and feedforward modules for domain adaptation.

In contrast, LoRA~\cite{TL:LoRA} stands out as a popular strategy with a re-parameterization property, allowing seamless integration into pre-trained models and bringing no extra computational costs during inference. 
Inspire by this, the subsequent research has introduced advanced techniques like factorized bilinear matrices~\cite{TL:FacT}, compactor decomposition~\cite{TL:Compacter}, Kronecker products~\cite{edalati2022krona}, and tensor factorization~\cite{chen2024superlora} to further enhance adaptation capabilities. Other studies~\cite{chavan2023one,wang2023multilora} explore multiple parallel layers to enhance the transfer representations. However, they suffer from two key issues: (1) how to create more compact and diverse feature spaces for complex domain adaptation, and (2) how to maintain consistency in intermediate distributions with pre-trained parameters.

To address this, we introduce a novel approach called Multi-Kernel Kronecker Adaptation with Re-Scaling Transmission (KARST) for visual classification. Specifically, our KARST leverages adaptive and flexible combinations to create compact mapping subspaces and build powerful feature patterns. Additionally, learnable re-scaling factors are employed to align diverse feature representations with the feature distributions of pre-trained parameters. Extensive experiments validate that our KARST not only outperforms existing PETL approaches and fully fine-tuning strategy, but also exhibits strong capability across different network backbones with minimal inference costs due to its re-parameterization characteristics.

\section{METHODOLOGY}
\subsection{Preliminary}
\label{AA}

Mathematically, the Kronecker product~\cite{edalati2022krona} extends the outer product from vectors to matrices and represents the tensor product in the standard basis, denoted by the symbol $\otimes$. 
Given the matrix $\mathbf{A} \in \mathbbm{R}^{p_{1}\times q_{1}}$ and the matrix $\mathbf{B} \in \mathbbm{R}^{p_{2} \times q_{2}}$, it transforms $\mathbf{A}\otimes \mathbf{B}$ into a new matrix with $(p_{1} \cdot p_{2})\times(q_{1} \cdot q_{2})$. Note that each element of the Kronecker product $\mathbf{A}\otimes \mathbf{B}$ can be represented as follows:

\begin{equation}
\label{eq:kronA}
\begin{pmatrix}
a_{11}\mathbf{B} & a_{12}\mathbf{B} & \cdots & a_{1q_{1}}\mathbf{B} \\
a_{21}\mathbf{B} & a_{22}\mathbf{B} & \cdots & a_{2q_{1}}\mathbf{B} \\
\vdots & \vdots & \ddots & \vdots \\
a_{p_{1}1}\mathbf{B} & a_{p_{1}2}\mathbf{B} & \cdots & a_{p_{1}q_{1}}\mathbf{B}
\end{pmatrix}_{(p_{1}\cdot p_{2})\times(q_{1} \cdot q_{2})}
\end{equation}
where $a_{p_{1}q_{1}}$ denotes the element in the $p_{1}$-th row and $q_{1}$-th column of the matrix $\mathbf{A}$.

\subsection{Multi-Kernel Kronecker Adaptation}
\begin{figure}[htp]
    \centering
    \includegraphics[width=\linewidth]{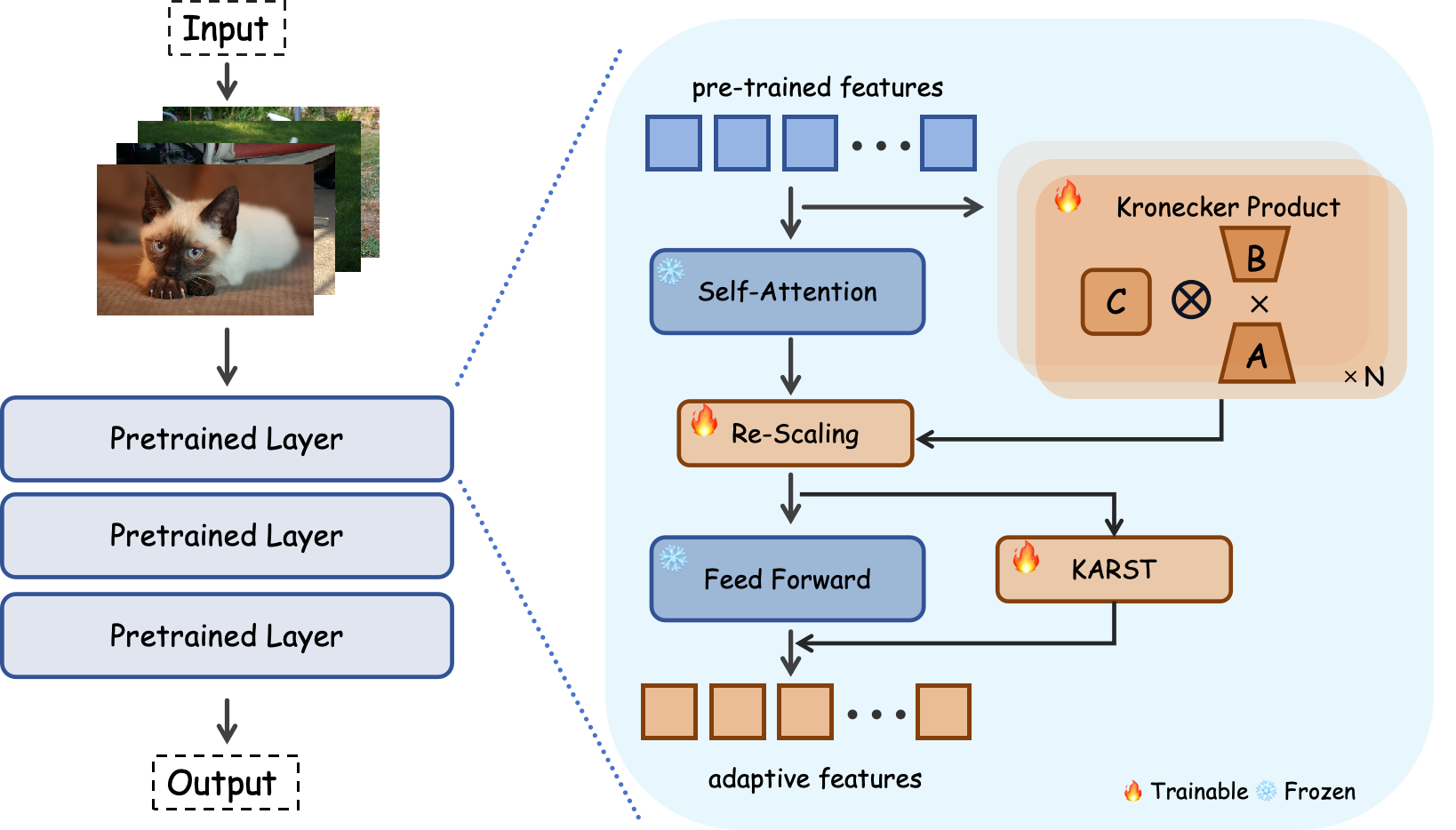}
    \caption{Framework of our proposed KARST. We first transform hidden features into $N$ kernel Kronecker products and then utilize shifting and scaling factors to align the merged outputs with the subsequent pre-trained layer.}
    \label{fig:framework}
\end{figure}

The parameters of the pre-trained ViT and final model are defined as $\mathbf{W_0}$ and $\mathbf{W_t}$. Here $\Delta{\mathbf{W}} = \mathbf{W_t} - \mathbf{W_0} $ denotes the updated weight during the fine-tuning process. Existing works attempt to decompose $\Delta{\mathbf{W}}$ into different forms to reduce the number of trainable parameters. For example, LoRA~\cite{TL:LoRA} decompose the $\Delta{\mathbf{W}}$ into two matrics $\mathbf{A}\in \mathbbm{R}^{D_{in} \times r}$ and $\mathbf{B} \in \mathbbm{R}^{r \times D_{out}}$ with the rank of r, while the Kronecker product~\cite{edalati2022krona} utilizes Eq.~\eqref{eq:kronA} as $\Delta{\mathbf{W}}$ to further enhance the adaptation capability. 
However, it increases the number of learnable parameters and creates only a single subspace with poor capabilities for diverse and complex data distributions. 
To address this, we propose multi-kernel Kronecker products in Fig.~\ref{fig:framework} with multiple complementary subspaces, which explore the limit of single mapping space and implicitly foster more adaptive representations across the subspaces.

We first decompose $\Delta{\mathbf{W}}$ into $N$ kernels of Kronecker space, which are defined as follows:
\begin{equation}
\label{eq:raw-multi-kernel}
    \Delta{\mathbf{W}} = \sum_{i=1}^{N} \mathbf{C}_{i} \otimes \mathbf{D}_{i} ,
\end{equation}
where $\mathbf{C}_{i} \in \mathbbm{R}^{m\times m}, \mathbf{D}_{i} \in \mathbbm{R}^{\frac{D_{in}}{m}\times \frac{D_{out}}{m}}$ are two decomposed matrics of $i$-th mapping space. Notably, $m$ is the stacking dimension of $\mathbf{C}_{i}$ and the resulting $\Delta{\mathbf{W}}$ is not
rank deficient~\cite{edalati2022krona}. 
Using multiple kernel functions can enrich the representation space by adaptively applying various mapping paradigms. During training, such an operation is performed in parallel with the pre-trained weight matrices. After fine-tuning, these factors can be merged back into the original weight matrix, ensuring no extra inference time.

However, this multi-kernel design inevitably introduces a large number of learnable parameters. To further reduce the computational burden, we decompose the matrix $\mathbf{D}_{i}$ into two sub-matrices $\mathbf{A}_{i}\in \mathbbm{R}^{\frac{D_{in}}{m} \times r}$ and $\mathbf{B}_{i} \in \mathbbm{R}^{r \times \frac{D_{out}}{m}}$. The Eq.~\eqref{eq:raw-multi-kernel} can be re-defined as follows:

\begin{equation}
\label{eq:new-multi-kernel}
    \Delta{\mathbf{W}} = \sum_{i=1}^{N} \mathbf{C}_{i} \otimes (\mathbf{A}_{i}\mathbf{B}_{i}) .
\end{equation}
Notably, we use random Gaussian initialization for the matrics $\mathbf{A}_{i}, \mathbf{C}_{i}$ and zero initialization for the matrix $\mathbf{B}_{i}$. This ensures that
$\Delta{\mathbf{W}}$ is zero at the beginning of training stage, so the initial state of the transformer block remains unchanged. As a result, our module can be inserted into any part of a pre-trained backbone without disrupting its initial status.

\subsection{Re-Scaling Transmission}
Existing PETL strategies~\cite{TL:FacT,TL:AdaLoRA} typically freeze pre-trained weights during fine-tuning to avoid overfitting and preserve the generalized patterns learned during the pre-training phase. However, the updated weights alert the value distribution of intermediate features compared to the pre-training phase. 
The misalignment between pre-trained layers and their corresponding input space limits the model's ability to adapt effectively to specific downstream tasks. Therefore, it is crucial to adjust the merging features adaptively to better align with the pre-trained mapping subspaces.

To address this challenge, we introduce a simple yet effective strategy to adjust the merged features from the pre-trained and Kronecker layers using two additional re-scaling parameters. Specifically, the feature representations are processed by performing a dot product with a scale factor and then adding a shift factor. Both of these parameters are independent of the input features, ensuring that the feature distribution can be effectively modified to align with the downstream dataset. This can be formulated as follows:
\begin{equation}
\mathbf{y} = (\mathbf{s_{1}} + \mathbbm{1}) \odot (\mathbf{W_0} + \Delta{\mathbf{W}}) \mathbf{x} + \mathbf{s_{2}}, 
\end{equation}
where $\mathbf{s_{1}}$ and $\mathbf{s_{2}}$ are channel-wise scaling and shift factors.
By incorporating the re-scaling transformation into the feature forward process, the network acquires more precise control over the feature distribution during fine-tuning, improving the network's ability to adapt to task-specific distributions.

Notably, we initialize $\mathbf{s_{1}}$ and $\mathbf{s_{2}}$ as zero vectors to ensure that the initial state of the pre-trained backbone remains unchanged. This preserves the learned representations of the pre-trained stage and prevents potentially unstable optimization of the model convergence during fine-tuning.
Since it is entirely linear, it displays an advantage of re-parameterization characteristics. This means that the channel-wise scaling and shifting operations can be effectively integrated into the preceding linear layers of the pre-trained network, thereby maintaining the simplicity and efficiency of the model architecture.

\begin{table*}[!htbp]
\caption{Results on the VTAB-1K benchmark with ViT-B/16. The top two best results are highlighted in bold and underlined mark.}
\centering
\setlength{\tabcolsep}{0.3pt}
\small
\begin{tabular}{p{2.2cm}<{}p{0.65cm}<{\centering}|p{0.65cm}<{\centering}p{0.65cm}<{\centering}p{0.65cm}<{\centering}p{0.65cm}<{\centering}p{0.65cm}<{\centering}p{0.65cm}<{\centering}p{0.65cm}<{\centering}|p{0.65cm}<{\centering}p{0.65cm}<{\centering}p{0.65cm}<{\centering}p{0.65cm}<{\centering}|p{0.65cm}<{\centering}p{0.65cm}<{\centering}p{0.65cm}<{\centering}p{0.65cm}<{\centering}p{0.65cm}<{\centering}p{0.65cm}<{\centering}p{0.65cm}<{\centering}p{0.65cm}<{\centering}|p{0.65cm}<{\centering}p{0.65cm}<{\centering}}
	\toprule[1.5pt]
	\multicolumn{2}{c|}{}&\multicolumn{7}{c|}{\textbf{Natural}}&\multicolumn{4}{c|}{\textbf{Specialized}}&\multicolumn{8}{c|}{\textbf{Structured}}&\\
	&\multicolumn{1}{c|}{\STAB{\rotatebox[origin=c]{90}{param (M)}}}
	&\multicolumn{1}{c}{\STAB{\rotatebox[origin=c]{90}{Cifar100}}}
	&\multicolumn{1}{c}{\STAB{\rotatebox[origin=c]{90}{Caltech101}}}
	&\multicolumn{1}{c}{\STAB{\rotatebox[origin=c]{90}{DTD}}}
	&\multicolumn{1}{c}{\STAB{\rotatebox[origin=c]{90}{Flower102}}}
	&\multicolumn{1}{c}{\STAB{\rotatebox[origin=c]{90}{Pets}}}
	&\multicolumn{1}{c}{\STAB{\rotatebox[origin=c]{90}{SVHN}}}
	&\multicolumn{1}{c|}{\STAB{\rotatebox[origin=c]{90}{Sun397}}}
	&\multicolumn{1}{c}{\STAB{\rotatebox[origin=c]{90}{Camelyon}}}
	&\multicolumn{1}{c}{\STAB{\rotatebox[origin=c]{90}{EuroSAT}}}
	&\multicolumn{1}{c}{\STAB{\rotatebox[origin=c]{90}{Resisc45}}}
	&\multicolumn{1}{c|}{\STAB{\rotatebox[origin=c]{90}{Retinopathy}}}
	&\multicolumn{1}{c}{\STAB{\rotatebox[origin=c]{90}{Clevr-Count}}}
	&\multicolumn{1}{c}{\STAB{\rotatebox[origin=c]{90}{Clevr-Dist}}}
	&\multicolumn{1}{c}{\STAB{\rotatebox[origin=c]{90}{DMLab}}}
	&\multicolumn{1}{c}{\STAB{\rotatebox[origin=c]{90}{KITTI-Dist}}}
	&\multicolumn{1}{c}{\STAB{\rotatebox[origin=c]{90}{dSpr-Loc}}}
	&\multicolumn{1}{c}{\STAB{\rotatebox[origin=c]{90}{dSpr-Ori}}}
	&\multicolumn{1}{c}{\STAB{\rotatebox[origin=c]{90}{sNORB-Azim}}}
	&\multicolumn{1}{c|}{\STAB{\rotatebox[origin=c]{90}{sNORB-Ele}}}
        &\multicolumn{1}{c}{\STAB{\rotatebox[origin=c]{90}{Average}}}
	&\multicolumn{1}{c}{\STAB{\rotatebox[origin=c]{90}{All Set Average}}}\\
	\specialrule{0em}{1pt}{1pt}
	\hline
	\specialrule{0em}{1pt}{1pt}
	\multicolumn{22}{l}{\emph{Traditional Fine-Tuning}}\\
	\hline
	\specialrule{0em}{1pt}{1pt}
	Full&85.8&68.9&87.7&64.3&97.2&86.9&87.4&38.8&79.7&95.7&84.2&73.9&56.3&58.6&41.7&65.5&57.5&46.7&25.7&29.1&68.9 &65.6 \\
	Linear&0&64.4&85.0&63.2&97.0&86.3&36.6&51.0&78.5&87.5&68.5&74.0&34.3&30.6&33.2&55.4&12.5&20.0&9.6&19.2&57.6 & 53.0\\
	\hline
	\specialrule{0em}{1pt}{1pt}
	\multicolumn{22}{l}{\emph{PETL methods}}\\
	\hline
	\specialrule{0em}{1pt}{1pt}
	BitFit&0.10&72.8&87.0&59.2&97.5&85.3&59.9&51.4&78.7&91.6&72.9&69.8&61.5&55.6&32.4&55.9&66.6&40.0&15.7&25.1&65.2 &62.0\\
    VPT&0.56&\textbf{78.8}&90.8&65.8&98.0&88.3&78.1&49.6&81.8&96.1&83.4&68.4&68.5&60.0&46.5&72.8&73.6&47.9&32.9&37.8&72.0 & 69.4 \\
	LST&2.38&59.5&91.5&69.0&99.2&89.9&79.5&54.6&86.9&95.9&85.3&74.1&81.8&61.8&52.2&81.0&71.7&49.5&33.7&45.2&74.3 &71.7\\
	LoRA&0.29&67.1&91.4&69.4&98.8&90.4&85.3&54.0&84.9&95.3&84.4&73.6&82.9&\textbf{69.2}&49.8&78.5&75.7&47.1&31.0&44.0&74.5 &72.3\\
	AdaptFormer&0.16&70.8&91.2&70.5&99.1&90.9&86.6&54.8&83.0&95.8&84.4&\underline{76.3}&81.9&64.3&49.3&80.3&76.3&45.7&31.7&41.1&74.7 &72.3 \\
    NOAH&0.43&69.6&92.7&70.2&99.1&90.4&86.1&53.7&84.4&95.4&83.9&75.8&82.8&68.9&49.9&81.7&81.8&48.3&32.8&44.2&75.5 & 73.2\\
	FacT&0.07&70.6&90.6&70.8&99.1&90.7&88.6&54.1&84.8&96.2&84.5&75.7&82.6&68.2&49.8&80.7&80.8&47.4&33.2&43.0&75.6 & 73.2\\
	SSF&0.21&69.0&92.6&\underline{75.1}&99.4&91.8&90.2&52.9&87.4&95.9&87.4&75.5&75.9&62.3&\textbf{53.3}&80.6&77.3&\textbf{54.9}&29.5&37.9&75.7 &73.2 \\

    DTL&0.04&69.6&\underline{94.8}&71.3&99.3&91.3&83.3&56.2&87.1&96.2&86.1&75.0&82.8&64.2&48.8&81.9&\textbf{93.9}&\underline{53.9}&34.2&\textbf{47.1}&76.7 &74.6 \\

    HEAT&-&72.0&92.3&71.4&99.2&91.4&90.2&55.9&88.0&95.8&85.5&75.5&\underline{83.7}&64.9&52.3&82.3&\underline{86.7}&53.5&\textbf{40.0}&44.8& 77.2 & 75.0 \\
    GLoRA&0.29&76.1&92.7&\textbf{75.3}&\textbf{99.6}&\textbf{92.4}&\underline{90.5}&57.2&87.5&\textbf{96.7}&\underline{88.1}&76.1&81.0&66.2&52.4&\textbf{84.9}&81.8&53.3&33.3&39.8&77.3 &75.0 \\
    
    Sparse-Tuning&0.32&74.8&\textbf{95.5}&73.2&99.4&91.7&88.1&\textbf{58.7}&\underline{88.2}&\underline{96.4}&85.8&\textbf{76.4}&82.9&64.7&50.7&\underline{83.4}&83.9&53.7&\underline{35.2}&45.2& 77.9 & 75.2 \\

    \hline
    \specialrule{0em}{1pt}{1pt}
	KARST($r\leq{8}$)&0.33&\underline{76.8}&93.2&\underline{75.1}&\underline{99.5}&\underline{92.2}&\textbf{91.9}&\underline{57.6}&\textbf{88.3}&96.2&\textbf{88.4}&75.7&\textbf{83.8}&\underline{69.0}&\underline{52.9}&82.0&86.0&52.9&33.8&\underline{47.0}& \textbf{78.1} & \textbf{75.9}\\
	\bottomrule[1.5pt]
	\end{tabular}
 	\label{tab:vtab-vit}
\end{table*}

\section{Experiments}
\subsection{Dataset Descriptions}
We conduct comprehensive experiments to prove KARST's effectiveness on VTAB-1K~\cite{zhai2019large} and few-shot datasets:

\textbf{VTAB-1k Benchmark}. It consists of 19 visual classification datasets. \textit{(1) Natural group} with classical vision problems contains natural images captured with standard cameras, including Caltech101~\cite{Datasets:Caltech101}, CIFAR100~\cite{Datasets:Cifar100}, DTD~\cite{Datasets:DTD}, Flowers102~\cite{Datasets:Flowers102}, Pets~\cite{Datasets:Pets}, Sun397~\cite{Datasets:SUN}, and SVHN~\cite{Datasets:SVHN}; \textit{(2) Specialized group} contains images captured from remote-sensing and medical device, including Resisc45~\cite{Datasets:Resisc}, EuroSAT~\cite{Datasets:Eurosat}, Patch Camelyon~\cite{Datasets:PatchCamelyon}, and Diabetic Retinopathy~\cite{Datasets:Retinopathy}; \textit{(3) Structured group} reports structured comprehension of a scene, e.g. object counting or 3D depth prediction, including Clevr~\cite{Datasets:Clevr}, dSprites~\cite{Datasets:DSprites}, SmallNORB~\cite{Datasets:SmallNORB}, DMLab~\cite{Datasets:Dmlab}, and KITTI~\cite{Datasets:Kitti}.

\textbf{Few-shot Benchmark}. We further conduct experiments on five standard fine-grained datasets under few-shot settings, including FGVC-Aircraft~\cite{maji2013fine}, Oxford-Pets~\cite{Datasets:Pets}, Food-101~\cite{bossard2014food}, Stanford Cars~\cite{krause20133d}, and Oxford-Flowers102~\cite{1640927}. These datasets encompass fine-grained classes from five categories in real-world scenarios: aircrafts, pets, food, cars, and flowers. Following previous work~\cite{TL:NOAH}, we evaluate various methods in {1, 2, 4, 8, 16}-shot settings for comparison.

\begin{figure*}[htp]
    \centering
    \includegraphics[width=1.0\linewidth]{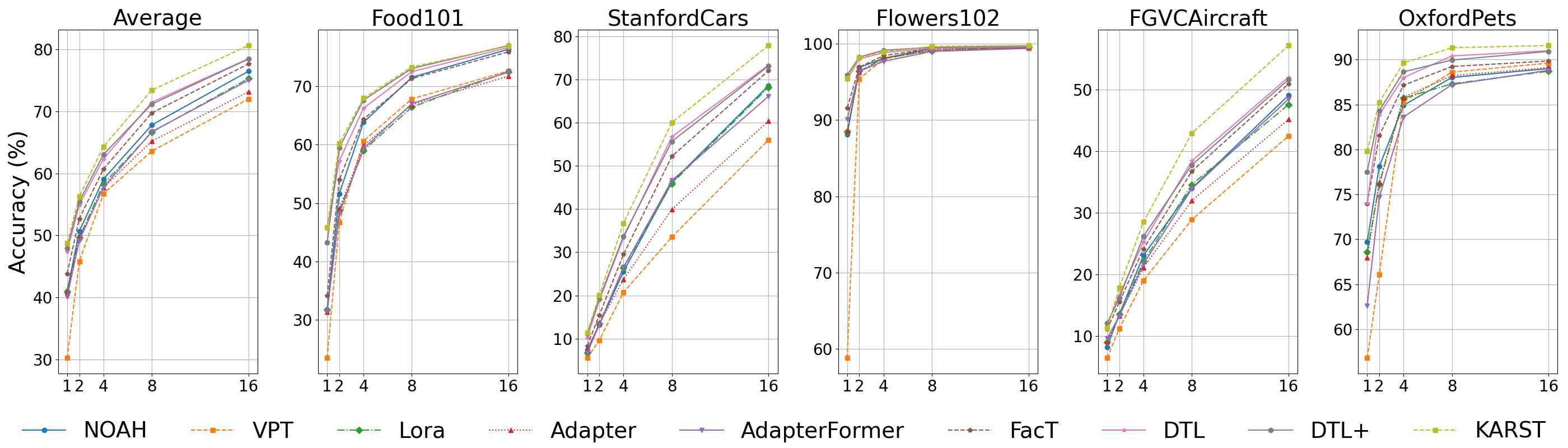}
    \caption{Top-1 accuracy on fine-grained few-shot benchmark with ViT-B/16 as the backbone. Note that our KARST significantly outperforms other PETL competitors and consistently achieves the best results under the few-shot settings across the five fine-grained visual classification datasets.}
    \label{fig:few-shot}
\end{figure*}

\subsection{Experiments on VTAB-1K Benchmark}

In this section, we select the vision transformer~\cite{TransF:ViT} (ViT-B/16) pre-trained on the ImageNet-21K~\cite{recht2019imagenet} dataset as the baseline unless otherwise specified. Besides, we also evaluate KARST on the Swin Transformer~\cite{TransF:SwinTransformer} (Swin-B) to validate its application capability across diverse architectures.

\begin{table}[htbp]
\caption{Results on the VTAB-1K benchmark with Swin-B.}
	\centering
	\small
	\setlength{\tabcolsep}{1pt}
	\begin{tabular}{p{1.5cm}<{}p{1.5cm}<{\centering}p{1cm}<{\centering}p{1cm}<{\centering}p{1cm}<{\centering}p{1cm}<{\centering}}
	\toprule[1.5pt]
	Method &param(M)&Nat.&Spe.&Str.&Avg.\\\hline
	\specialrule{0em}{1pt}{1pt}
	Full&86.7 &79.2&86.2&59.7&75.0\\
	Linear&0&73.5&80.8&33.5&62.6\\
	BitFit&0.20&74.2&80.1&42.4&65.6\\
	VPT& 0.16&76.8&84.5&53.4&71.6\\
        HEAT & - & 82.3 & \underline{87.1} & 61.4 & 76.9 \\
	FacT&0.14&\underline{83.1}&86.9&\underline{62.1}&77.4\\
 	DTL&0.09&82.4&87.0&\textbf{64.2}&77.9\\
	KARST&0.45&\textbf{83.9}&\textbf{87.7}&\textbf{64.2}&\textbf{78.6}\\
	\bottomrule[1.5pt]
	\end{tabular}
	\label{tab:vtab-swin}
\end{table}

\textbf{Baselines.} Using ViT-B/16 and Swin-B models, we compare our KARST with fully fine-tuning (Full), partially-tuning with specific task head (Linear), and several competitive strategies, including {BitFit}~\cite{TL:BitFit}, {VPT}~\cite{TL:VPT}, {LST}~\cite{TL:LST}, {LoRA}~\cite{TL:LoRA}, {AdaptFormer}~\cite{TL:AdaptFormer}, {NOAH}~\cite{TL:NOAH}, {FacT}~\cite{TL:FacT}, {SSF}~\cite{TL:SSF}, {DTL}~\cite{fu2024dtl}, {HEAT}~\cite{TL:heat}, {GLoRA}~\cite{chavan2023one}, and {Sparse-Tuning}~\cite{TL:Sparse-Tuning}. Note that we set $r\leq32$ for FacT, $r=0.9$ for Sparse-Tuning, and $r\leq 8$ for LoRA and KARST. The prompt length $l$ for VPT is used from the original paper. Besides, the stacking dimension $m$ and kernel number $N$ are 8 and 2, respectively.

\textbf{Results.} The results using ViT-B and Swin-B are summarized in TABLE~\ref{tab:vtab-vit} and TABLE~\ref{tab:vtab-swin}, respectively. We can discover that KARST produces the best performance with similar trainable parameter usage compared to existing state-of-the-art PETL methods. Notably, KARST can achieve 78.1$\%$ and 78.6$\%$ across different network architectures, surpassing the best competitors Sparse-Tuning and DTL on the 19 datasets.

\subsection{Experiments on Few-shot Benchmark}
\textbf{Baselines.} For few-shot learning, we employ the ViT-B/16 model pre-trained on ImageNet-21K as the baseline. We compare our KARST with {LoRA}, {FacT}, {NOAH}, {Adapter}, {AdapterFormer}, {VPT} and {DTL} across five fine-grained datasets. We maintain their official hyper-parameter configurations and report the averaging results over three runs with different random seeds. Note that we set the rank $r\leq 8$, the stacking dimension $m=8$, and kernel number $N=2$, respectively.

\textbf{Results.} 
From Fig.~\ref{fig:few-shot}, we can observe that our proposed KARST consistently achieves the best performance across five fine-grained datasets. Notably, KARST significantly outperforms existing PETL strategies by a large margin in few-shot settings, highlighting its superior effectiveness and generalizability, particularly when limited data is available. This verifies the adaptability and capability of KARST to extract meaningful features even with minimal training samples.
\subsection{Ablation Study}

\begin{table}
	\caption{Average accuracy with or without Re-Scaling Transmission. We utilize ViT-B/16 as the baseline on VTAB-1K benchmark.}
	\centering
	\small
	\setlength{\tabcolsep}{1pt}
	\begin{tabular}{p{1.4cm}<{}p{1.4cm}<{\centering}p{0.7cm}<{\centering}p{0.7cm}<{\centering}p{0.7cm}<{\centering}p{0.7cm}<{\centering}p{1.4cm}<{\centering}}
	\toprule[1.5pt]
	Method &param (M)&Nat.&Spe.&Str.&Avg.& Set Avg.\\\hline
	\specialrule{0em}{1pt}{1pt}
        FacT & 0.07 &80.6&85.3&60.7&75.6 & 73.2\\
 	+ RST & 0.07 &\textbf{81.9}&\textbf{85.7}&\textbf{61.2}&\textbf{76.3} & \textbf{74.0} \\
        KA & 0.30 &83.2&86.5&63.0& 77.5 & 75.4 \\
	+ RST &0.33 & \textbf{83.7} & \textbf{87.2} & \textbf{63.4} & \textbf{78.1} &\textbf{75.9} \\

	\bottomrule[1.5pt]
	\end{tabular}
	\label{tab:ablation-rst}
\end{table}

\textbf{Effectiveness of Re-Scaling Transmission}. In TABLE~\ref{tab:ablation-rst}, we evaluate the role of the re-scaling module. KA and RST denote multi-kernel Kronecker Adaptation and re-scaling transmission. We notice that averaging performance gains are 0.7\% and 0.6\% when our RST module works with FacT and our KA module. This shows our RST module can effectively mitigate misalignment issues inside PETL methods and transform intermediate features to better align with pre-trained patterns. The valid and consistent improvements validate the robustness and general applicability of our RST module.

\begin{figure}[t]
    \centering
    \includegraphics[width=0.95\linewidth,trim= 20 30 20 27,clip]{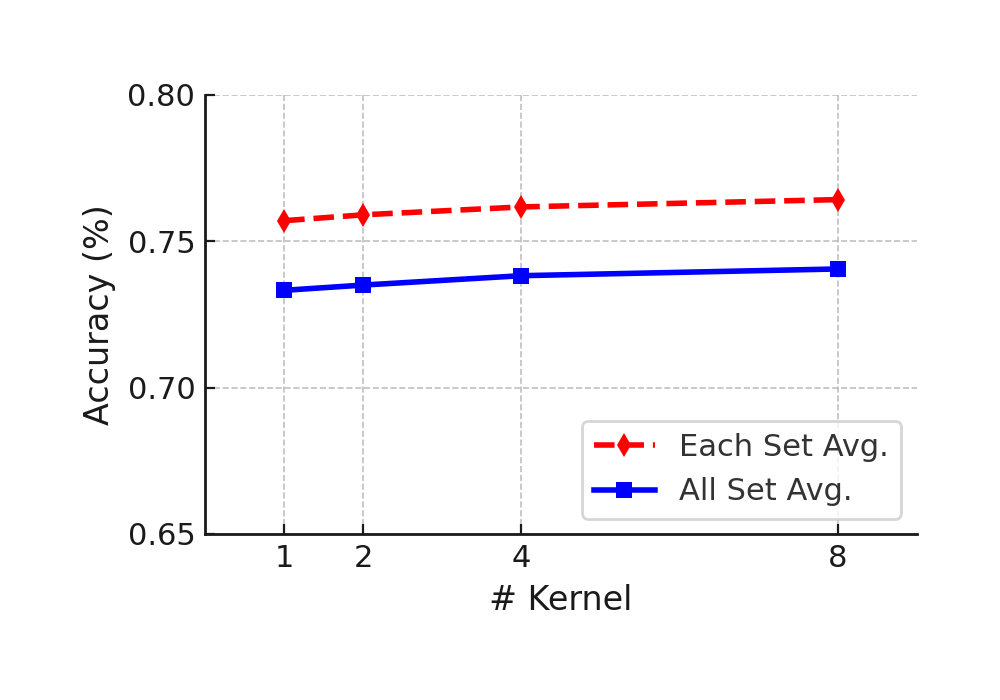}
    \caption{Average accuracy of KARST on VTAB-1K with multiple kernels.}
    \label{fig:multi-group}
\end{figure}

\textbf{Hyper-parameter of the kernel number}. Fig.~\ref{fig:multi-group} demonstrate the influence of different kernel numbers for multi-kernel Kronecker Adaptation. The ViT-B/16 is adopted as the baseline on the VTAB-1K benchmark. 
We observe that as the number of kernels increases, the average accuracy improves. However, this also results in a corresponding rise in the number of fine-tuned parameters. After careful evaluation of the trade-off between performance and parameter efficiency, we selected a kernel number of 2 for our experiments.

\section{Conclusion}

In this paper, we propose a novel approach called Multi-Kernel Kronecker Adaptation with Re-Scaling Transmission (KARST) for recognition tasks. Specifically, KARST leverages multiple kernel functions within Kronecker products to build compact mapping spaces, enhancing the model's representation capacity for robust adaptation. Additionally, a simple yet effective re-scaling strategy is employed to align the resulting features with the patterns and distribution of the pre-trained weights. Extensive experiments across 24 diverse datasets demonstrate the general effectiveness and broad applicability of KARST, which can also cooperate with existing PETL techniques to achieve better performance gains.

\bibliographystyle{IEEEtran}
\bibliography{strings}

\begin{thebibliography}{10}
\providecommand{\url}[1]{#1}
\csname url@samestyle\endcsname
\providecommand{\newblock}{\relax}
\providecommand{\bibinfo}[2]{#2}
\providecommand{\BIBentrySTDinterwordspacing}{\spaceskip=0pt\relax}
\providecommand{\BIBentryALTinterwordstretchfactor}{4}
\providecommand{\BIBentryALTinterwordspacing}{\spaceskip=\fontdimen2\font plus
\BIBentryALTinterwordstretchfactor\fontdimen3\font minus \fontdimen4\font\relax}
\providecommand{\BIBforeignlanguage}[2]{{%
\expandafter\ifx\csname l@#1\endcsname\relax
\typeout{** WARNING: IEEEtran.bst: No hyphenation pattern has been}%
\typeout{** loaded for the language `#1'. Using the pattern for}%
\typeout{** the default language instead.}%
\else
\language=\csname l@#1\endcsname
\fi
#2}}
\providecommand{\BIBdecl}{\relax}
\BIBdecl

\bibitem{TransF:ViT}
A.~Dosovitskiy, L.~Beyer, A.~Kolesnikov, D.~Weissenborn, X.~Zhai, T.~Unterthiner, M.~Dehghani, M.~Minderer, G.~Heigold, S.~Gelly, J.~Uszkoreit, and N.~Houlsby, ``An image is worth 16x16 words: Transformers for image recognition at scale,'' in \emph{ICLR}, 2021.

\bibitem{VLM:EVE}
H.~Diao, Y.~Cui, X.~Li, Y.~Wang, H.~Lu, and X.~Wang, ``Unveiling encoder-free vision-language models,'' \emph{arXiv:2406.11832}, 2024.

\bibitem{ITM:SGRAF}
H.~Diao, Y.~Zhang, L.~Ma, and H.~Lu, ``Similarity reasoning and filtration for image-text matching,'' in \emph{AAAI}, 2021, pp. 1218--1226.

\bibitem{TL:UniPT}
H.~Diao, B.~Wan, Y.~Zhang, X.~Jia, H.~Lu, and L.~Chen, ``Unipt: Universal parallel tuning for transfer learning with efficient parameter and memory,'' in \emph{CVPR}, 2024.

\bibitem{diao2025sherl}
H.~Diao, B.~Wan, X.~Jia, Y.~Zhuge, Y.~Zhang, H.~Lu, and L.~Chen, ``Sherl: Synthesizing high accuracy and efficient memory for resource-limited transfer learning,'' in \emph{ECCV}, 2025, pp. 75--95.

\bibitem{TL:BitFit}
E.~B. Zaken, Y.~Goldberg, and S.~Ravfogel, ``Bitfit: Simple parameter-efficient fine-tuning for transformer-based masked language-models,'' in \emph{ACL}, 2022, pp. 1--9.

\bibitem{TL:VPT}
M.~Jia, L.~Tang, B.~Chen, C.~Cardie, S.~J. Belongie, B.~Hariharan, and S.~Lim, ``Visual prompt tuning,'' in \emph{ECCV}, vol. 13693, 2022, pp. 709--727.

\bibitem{diao2024gssf}
H.~Diao, Y.~Zhang, S.~Gao, J.~Zhu, L.~Chen, and H.~Lu, ``Gssf: Generalized structural sparse function for deep cross-modal metric learning,'' \emph{IEEE TIP}, 2024.

\bibitem{TL:LoRA}
E.~J. Hu, Y.~Shen, P.~Wallis, Z.~Allen{-}Zhu, Y.~Li, S.~Wang, L.~Wang, and W.~Chen, ``Lora: Low-rank adaptation of large language models,'' in \emph{ICLR}, 2022.

\bibitem{TL:FacT}
S.~Jie and Z.~Deng, ``Fact: Factor-tuning for lightweight adaptation on vision transformer,'' in \emph{AAAI}, 2023.

\bibitem{TL:Compacter}
R.~K. Mahabadi, J.~Henderson, and S.~Ruder, ``Compacter: Efficient low-rank hypercomplex adapter layers,'' in \emph{NeurIPS}, 2021, pp. 1022--1035.

\bibitem{edalati2022krona}
A.~Edalati, M.~Tahaei, I.~Kobyzev, V.~P. Nia, J.~J. Clark, and M.~Rezagholizadeh, ``Krona: Parameter efficient tuning with kronecker adapter,'' \emph{arXiv preprint arXiv:2212.10650}, 2022.

\bibitem{chen2024superlora}
X.~Chen, J.~Liu, Y.~Wang, M.~Brand, G.~Wang, T.~Koike-Akino \emph{et~al.}, ``Superlora: Parameter-efficient unified adaptation of multi-layer attention modules,'' \emph{arXiv preprint arXiv:2403.11887}, 2024.

\bibitem{chavan2023one}
A.~Chavan, Z.~Liu, D.~Gupta, E.~Xing, and Z.~Shen, ``One-for-all: Generalized lora for parameter-efficient fine-tuning,'' \emph{arXiv preprint arXiv:2306.07967}, 2023.

\bibitem{wang2023multilora}
Y.~Wang, Y.~Lin, X.~Zeng, and G.~Zhang, ``Multilora: Democratizing lora for better multi-task learning,'' \emph{arXiv preprint arXiv:2311.11501}, 2023.

\bibitem{TL:AdaLoRA}
Q.~Zhang, M.~Chen, A.~Bukharin, P.~He, Y.~Cheng, W.~Chen, and T.~Zhao, ``Adaptive budget allocation for parameter-efficient fine-tuning,'' in \emph{ICLR}, 2023.

\bibitem{zhai2019large}
X.~Zhai, J.~Puigcerver, A.~Kolesnikov, P.~Ruyssen, C.~Riquelme, M.~Lucic, J.~Djolonga, A.~S. Pinto, M.~Neumann, A.~Dosovitskiy \emph{et~al.}, ``A large-scale study of representation learning with the visual task adaptation benchmark,'' \emph{arXiv preprint arXiv:1910.04867}, 2019.

\bibitem{Datasets:Caltech101}
L.~Fei-Fei, R.~Fergus, and P.~Perona, ``One-shot learning of object categories,'' \emph{IEEE transactions on pattern analysis and machine intelligence}, vol.~28, no.~4, pp. 594--611, 2006.

\bibitem{Datasets:Cifar100}
A.~Krizhevsky, G.~Hinton \emph{et~al.}, ``Learning multiple layers of features from tiny images,'' 2009.

\bibitem{Datasets:DTD}
M.~Cimpoi, S.~Maji, I.~Kokkinos, S.~Mohamed, and A.~Vedaldi, ``Describing textures in the wild,'' in \emph{Proceedings of the IEEE conference on computer vision and pattern recognition}, 2014, pp. 3606--3613.

\bibitem{Datasets:Flowers102}
M.-E. Nilsback and A.~Zisserman, ``Automated flower classification over a large number of classes,'' in \emph{2008 Sixth Indian conference on computer vision, graphics \& image processing}.\hskip 1em plus 0.5em minus 0.4em\relax IEEE, 2008, pp. 722--729.

\bibitem{Datasets:Pets}
O.~M. Parkhi, A.~Vedaldi, A.~Zisserman, and C.~Jawahar, ``Cats and dogs,'' in \emph{2012 IEEE conference on computer vision and pattern recognition}.\hskip 1em plus 0.5em minus 0.4em\relax IEEE, 2012, pp. 3498--3505.

\bibitem{Datasets:SUN}
J.~Xiao, J.~Hays, K.~A. Ehinger, A.~Oliva, and A.~Torralba, ``Sun database: Large-scale scene recognition from abbey to zoo,'' in \emph{2010 IEEE computer society conference on computer vision and pattern recognition}.\hskip 1em plus 0.5em minus 0.4em\relax IEEE, 2010, pp. 3485--3492.

\bibitem{Datasets:SVHN}
Y.~Netzer, T.~Wang, A.~Coates, A.~Bissacco, B.~Wu, A.~Y. Ng \emph{et~al.}, ``Reading digits in natural images with unsupervised feature learning,'' in \emph{NIPS workshop on deep learning and unsupervised feature learning}, vol. 2011, no.~2.\hskip 1em plus 0.5em minus 0.4em\relax Granada, 2011, p.~4.

\bibitem{Datasets:Resisc}
G.~Cheng, J.~Han, and X.~Lu, ``Remote sensing image scene classification: Benchmark and state of the art,'' \emph{Proceedings of the IEEE}, vol. 105, no.~10, pp. 1865--1883, 2017.

\bibitem{Datasets:Eurosat}
P.~Helber, B.~Bischke, A.~Dengel, and D.~Borth, ``Eurosat: A novel dataset and deep learning benchmark for land use and land cover classification,'' \emph{IEEE Journal of Selected Topics in Applied Earth Observations and Remote Sensing}, vol.~12, no.~7, pp. 2217--2226, 2019.

\bibitem{Datasets:PatchCamelyon}
B.~S. Veeling, J.~Linmans, J.~Winkens, T.~Cohen, and M.~Welling, ``Rotation equivariant cnns for digital pathology,'' in \emph{Medical Image Computing and Computer Assisted Intervention--MICCAI 2018: 21st International Conference, Granada, Spain, September 16-20, 2018, Proceedings, Part II 11}.\hskip 1em plus 0.5em minus 0.4em\relax Springer, 2018, pp. 210--218.

\bibitem{Datasets:Retinopathy}
\BIBentryALTinterwordspacing
E.~Kaggle, ``Kaggle diabetic retinopathy detection,'' July 2015, accessed: Sep. 12, 2024. [Online]. Available: \url{https://www.kaggle.com/c/diabetic-retinopathy-detection}
\BIBentrySTDinterwordspacing

\bibitem{Datasets:Clevr}
J.~Johnson, B.~Hariharan, L.~Van Der~Maaten, L.~Fei-Fei, C.~Lawrence~Zitnick, and R.~Girshick, ``Clevr: A diagnostic dataset for compositional language and elementary visual reasoning,'' in \emph{Proceedings of the IEEE conference on computer vision and pattern recognition}, 2017, pp. 2901--2910.

\bibitem{Datasets:DSprites}
L.~Matthey, I.~Higgins, D.~Hassabis, and A.~Lerchner, ``dsprites: Disentanglement testing sprites dataset,'' 2017.

\bibitem{Datasets:SmallNORB}
Y.~LeCun, F.~J. Huang, and L.~Bottou, ``Learning methods for generic object recognition with invariance to pose and lighting,'' in \emph{Proceedings of the 2004 IEEE Computer Society Conference on Computer Vision and Pattern Recognition, 2004. CVPR 2004.}, vol.~2.\hskip 1em plus 0.5em minus 0.4em\relax IEEE, 2004, pp. II--104.

\bibitem{Datasets:Dmlab}
C.~Beattie, J.~Z. Leibo, D.~Teplyashin, T.~Ward, M.~Wainwright, H.~K{\"u}ttler, A.~Lefrancq, S.~Green, V.~Vald{\'e}s, A.~Sadik \emph{et~al.}, ``Deepmind lab,'' \emph{arXiv preprint arXiv:1612.03801}, 2016.

\bibitem{Datasets:Kitti}
A.~Geiger, P.~Lenz, C.~Stiller, and R.~Urtasun, ``Vision meets robotics: The kitti dataset. the international journal of robotics research,'' \emph{Int. J. Rob. Res}, pp. 1--6.

\bibitem{maji2013fine}
S.~Maji, E.~Rahtu, J.~Kannala, M.~Blaschko, and A.~Vedaldi, ``Fine-grained visual classification of aircraft,'' \emph{arXiv preprint arXiv:1306.5151}, 2013.

\bibitem{bossard2014food}
L.~Bossard, M.~Guillaumin, and L.~Van~Gool, ``Food-101--mining discriminative components with random forests,'' in \emph{Computer Vision--ECCV 2014: 13th European Conference, Zurich, Switzerland, September 6-12, 2014, Proceedings, Part VI 13}.\hskip 1em plus 0.5em minus 0.4em\relax Springer, 2014, pp. 446--461.

\bibitem{krause20133d}
J.~Krause, M.~Stark, J.~Deng, and L.~Fei-Fei, ``3d object representations for fine-grained categorization,'' in \emph{Proceedings of the IEEE international conference on computer vision workshops}, 2013, pp. 554--561.

\bibitem{1640927}
M.-E. Nilsback and A.~Zisserman, ``A visual vocabulary for flower classification,'' in \emph{2006 IEEE Computer Society Conference on Computer Vision and Pattern Recognition (CVPR'06)}, vol.~2, 2006, pp. 1447--1454.

\bibitem{TL:NOAH}
Y.~Zhang, K.~Zhou, and Z.~Liu, ``Neural prompt search,'' \emph{arXiv: 2206.04673}, 2022.

\bibitem{recht2019imagenet}
B.~Recht, R.~Roelofs, L.~Schmidt, and V.~Shankar, ``Do imagenet classifiers generalize to imagenet?'' in \emph{International conference on machine learning}.\hskip 1em plus 0.5em minus 0.4em\relax PMLR, 2019, pp. 5389--5400.

\bibitem{TransF:SwinTransformer}
Z.~Liu, Y.~Lin, Y.~Cao, H.~Hu, Y.~Wei, Z.~Zhang, S.~Lin, and B.~Guo, ``Swin transformer: Hierarchical vision transformer using shifted windows,'' in \emph{ICCV}.\hskip 1em plus 0.5em minus 0.4em\relax {IEEE}, 2021, pp. 9992--10\,002.

\bibitem{TL:LST}
Y.~Sung, J.~Cho, and M.~Bansal, ``{LST:} ladder side-tuning for parameter and memory efficient transfer learning,'' in \emph{NeurIPS}, 2022.

\bibitem{TL:AdaptFormer}
S.~Chen, C.~Ge, Z.~Tong, J.~Wang, Y.~Song, J.~Wang, and P.~Luo, ``Adaptformer: Adapting vision transformers for scalable visual recognition,'' in \emph{NeurIPS}, 2022.

\bibitem{TL:SSF}
D.~Lian, D.~Zhou, J.~Feng, and X.~Wang, ``Scaling {\&} shifting your features: {A} new baseline for efficient model tuning,'' in \emph{NeurIPS}, 2022.

\bibitem{fu2024dtl}
M.~Fu, K.~Zhu, and J.~Wu, ``Dtl: Disentangled transfer learning for visual recognition,'' in \emph{Proceedings of the AAAI Conference on Artificial Intelligence}, vol.~38, no.~11, 2024, pp. 12\,082--12\,090.

\bibitem{TL:heat}
Y.~Zhong and Y.~Zhou, ``Heat: Head-level parameter efficient adaptation of vision transformers with taylor-expansion importance scores,'' \emph{arXiv preprint arXiv:2404.08894}, 2024.

\bibitem{TL:Sparse-Tuning}
T.~Liu, X.~Liu, L.~Shi, Z.~Xu, S.~Huang, Y.~Xin, and Q.~Yin, ``Sparse-tuning: Adapting vision transformers with efficient fine-tuning and inference,'' \emph{arXiv preprint arXiv:2405.14700}, 2024.

\end{thebibliography}

\end{document}